\documentclass[runningheads]{llncs}
\usepackage[T1]{fontenc}
\usepackage{graphicx}
\usepackage{amssymb}
\usepackage{subcaption}

\begin{document}
\title{Histopathology image embedding based on foundation models features aggregation for patient treatment response prediction.}

\author{Bilel Guetarni\inst{1,2,3,4,5}(corresponding author: \email{bilel.guetarni@junia.com}) \and Feryal Windal\inst{1,2,3,4,5} \and Halim Benhabiles\inst{5,6} \and Mahfoud Chaibi\inst{7} \and Romain Dubois\inst{7} \and Emmanuelle Leteurtre\inst{7} \and Dominique Collard\inst{8,9}}
\authorrunning{B. Guetarni et al.}
\institute{Junia, Lille, France \and
UMR 8520, CNRS, France \and
Centrale Lille, France \and
Univerity of Polytechnique Hauts-de-France, Lille, France \and
University of Lille, France \and
Centre for Digital Systems, IMT Nord Europe, Institut Mines-Télécom, Lille, France \and
Department of Pathology of the University Hospital of Lille, France \and
LIMMS/CNRS-IIS, IRL 2820, The University of Tokyo, Japan \and
Yncrea Méditerranée, Yncrea Ouest, Junia, Toulon, Brest, Lille, France}
\maketitle              % typeset the header of the contribution
\begin{abstract}
Predicting the response of a patient to a cancer treatment is of high interest.
Nonetheless, this task is still challenging from a medical point of view due to the complexity of the interaction between the patient organism and the considered treatment.
Recent works on foundation models pre-trained with self-supervised learning on large-scale unlabeled histopathology datasets have opened a new direction towards the development of new methods for cancer diagnosis related tasks.
In this article, we propose a novel methodology for predicting Diffuse Large B-Cell Lymphoma patients treatment response from Whole Slide Images.
Our method exploits several foundation models as feature extractors to obtain a local representation of the image corresponding to a small region of the tissue, then, a global representation of the image is obtained by aggregating these local representations using attention-based Multiple Instance Learning.
Our experimental study conducted on a dataset of 152 patients, shows the promising results of our methodology, notably by highlighting the advantage of using foundation models compared to conventional ImageNet pre-training.
Moreover, the obtained results clearly demonstrates the potential of foundation models for characterizing histopathology images and generating more suited semantic representation for this task.
\end{abstract}

\section{Introduction}
\label{introduction}
Diffuse Large B-Cell Lymphoma (DLBCL) is the most common type of lymphoma (immune system cells cancer) \cite{Fan2017RituximabBasedTI} and approximately 40\% of DLBCL patients relapse following the initial immunochemotherapy \cite{Nowakowski2015LenalidomideCW}.
It has been shown that patients who failed to complete the treatment are associated with a low survival rate \cite{Wsterlid2020OutcomeAD}.
Therefore, identifying theses patients could allow to adapt their treatment and lead to a better outcome \cite{Coiffier2016DiffuseLB}.
In this sense, predicting the response to a treatment is of high interest, nonetheless, this task is still challenging from a medical point of view.
Currently, \textit{post-treatment} response of DLBCL cases is evaluated through the Deauville score \cite{Meignan2009ReportOT}, that permits to categorize patients into two classes, namely, positive response (i.e., complete response) and negative response (i.e., stable disease).
Hence, predicting this score before treatment initiation would allow to select the most promising treatment.

In this context, machine learning-based analysis methods have been investigated for several DLBCL diagnosis related tasks such as molecular subtyping \cite{Guetarni2023AVT}, tumor identification \cite{Li2020ADL} and survival rate prediction \cite{Carreras2021ArtificialNN}.

In the meantime, the recent emergence of foundation models, pre-trained through self-supervised learning (SSL), on large unlabeled Whole Slide Image (WSI) datasets, has permitted to improve the performance of pathology analysis tasks notably cancer diagnosis \cite{Xu2024AWF,Lu2024AVF,Filiot2023ScalingSL}.
Indeed, this new family of deep learning models has been shown to provides promising results for hitsopathology tasks such as cancer subtyping, genetic alterations or overall survival \cite{Filiot2023ScalingSL}.

In this article, we propose a novel methodology for predicting DLBCL treatment response from WSI.
More specifically, as illustrated in Fig. \ref{fig:architecture}, our method exploits several foundation models as patch-based feature extractors; these features are aggregated into one patch-based local representation, then, a global representation of the WSI is obtained by aggregating these local representations using attention-based Multiple Instance Learning (MIL).
The choice of this later technique is motivated by the fact that only WSI patches (instances) representing tumoral or stroma tissue are relevant for predicting the response.
The main contributions of our work are, (1) the proposition of a method for characterizing histopathology images by exploiting the aggregation of several foundation models offering a better semantic representation of the tissue compared to standard CNN models, (2) an experimental study conducted on a dataset of 152 patients, showing the promising results of our methodology and emphasizing the advantage of using foundation models compared to conventional ImageNet pre-trained models for the treatment response prediction task. Our code will be publicly available after the review process.

It is worth mentioning that very recently, Lee et al. \cite{Lee2024PredictionOI} have proposed a method for treatment response prediction from WSI using a Vision Transformer model exclusively trained on their own private dataset.
The main difference between this work and ours, is that our method produces a prediction model that is not only trained on our own dataset but also leverages representations learned from large-scale histopathology datasets through the use of foundation models.
We emphasizes the fact that we couldn't include this method in our experimental study since the dataset and the source code are not publicly available.
We also note that this prediction task has been also investigated with other types of features such as CT-based radiomics \cite{Santiago2021CTbasedRM} and protein expression markers \cite{Sez2004BuildingAO}.

\section{Methods}

\begin{figure}[h]
\centering
\includegraphics[trim={0cm 0cm 0cm 0cm},clip,width=\textwidth]{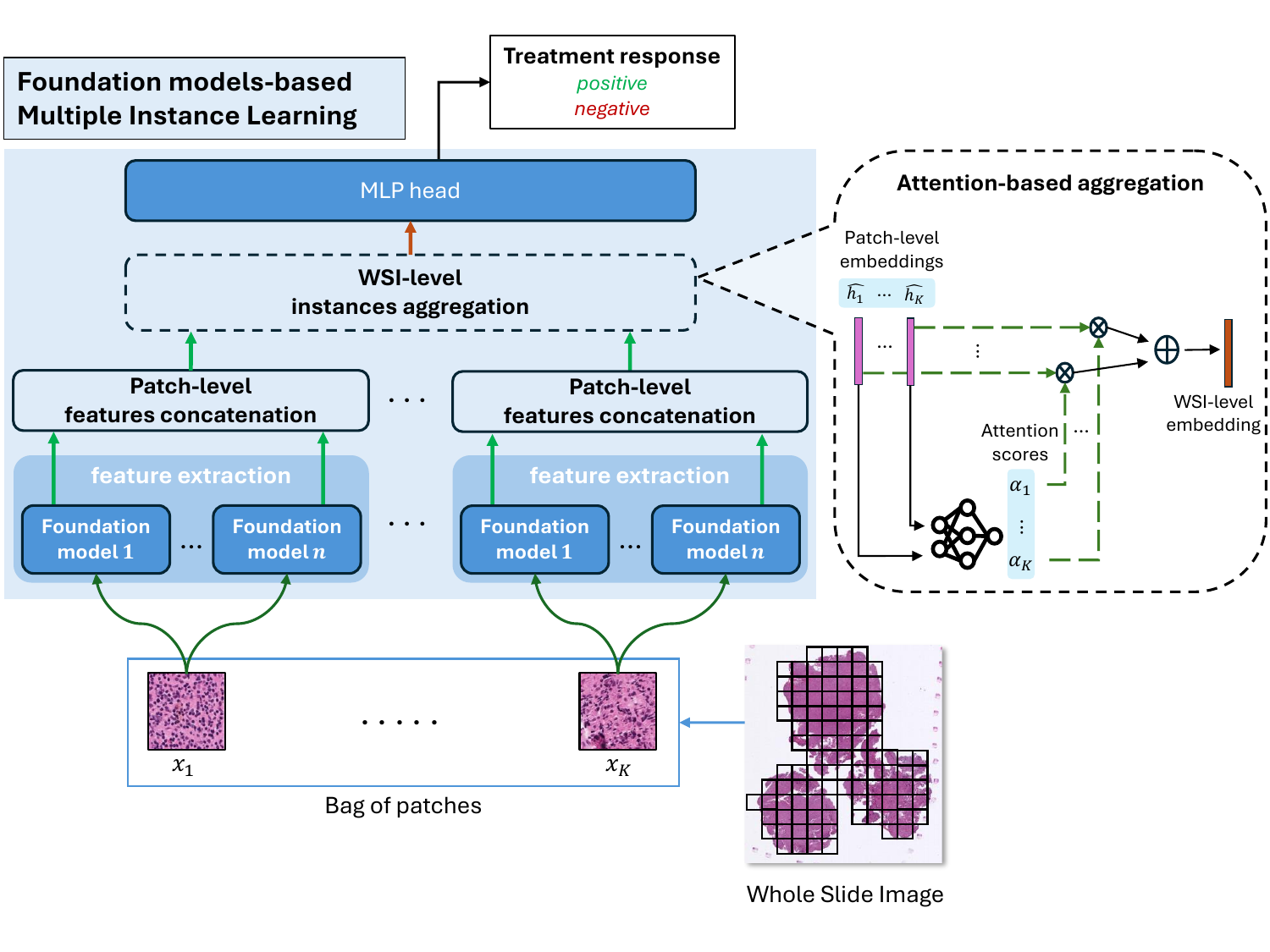}
\caption{WSI-based treatment response prediction for DLBCL patients with foundation models. Patch-level embeddings, extracted by multiple foundation models, are concatenated to create diverse and rich patch embeddings. An attention-based aggregation function is then applied to these, in order to obtain a WSI-level embedding. The treatment response is then predicted from this embedding.}
\label{fig:architecture}
\end{figure}

We consider the task of predicting the treatment response of a DLBCL patient based on HES stained WSI.
Indeed, WSI with HES staining is the standard histology support for cancer diagnosis, including forecasting therapy outcome, which makes such type of data usually available in medical centers.
These WSIs, are generally associated with the treatment response based on the patient's Deauville score \cite{Meignan2009ReportOT} (positive vs. negative response).
We are therefore considering a binary classification problem to build our prediction model.

Let's denote $X=\{x_{1}, ..., x_{K}\}$ a bag of patches representing a patient's WSI (see "Bag of patches" in Fig. \ref{fig:architecture}).
Each patch $x_{i} \in \mathbb{R}^{p \times p \times 3}$ is an RGB image of size $p \times p$.
Let's denote $y \in \{0,1\}$ the ground-truth label representing the treatment response associated to this WSI (1 for a positive response and 0 for a negative response).

We consider a set of $n$ foundation models, pre-trained on large-scale datasets of histopathology image, to be used as feature extractors (see "feature extraction" in Fig. \ref{fig:architecture}).
We denote this set as $\mathcal{F}=\{f^{1}, ..., f^{n}\}$.
Each feature extractor $f^{j}$ is applied to produce a low-dimensional embedding from each patch $x_{i}$, such as:
\begin{equation}
h_{i}^{j} = f^{j}(x_{i})
\end{equation}
Indeed, transforming patches into low-dimensional embedding using foundation models would permit to extract the most relevant histopathology features for tissue characterization, which is essential for treatment response prediction.
It is worth mentioning that most existing histopathology foundation models are based on Vision Transformer architectures \cite{Lu2024AVF,Xu2024AWF,Chen2022ScalingVT}.
For this reason, $h_{i}^{j}$ corresponds to the special output \textit{CLS} token of the model $f^{j}$.

Next, we define a patch-level aggregation function, denoted $\sigma$, that aggregates the different embeddings into a single representation of the patch (see "patch-level features concatenation" in Fig. \ref{fig:architecture}):
\begin{equation}
\hat{h}_{i} = \sigma(h_{i}^{1}, ..., h_{i}^{n})
\end{equation}
where $\sigma$ is defined as the concatenation operation.

We made the choice to use this operation to fully utilize the embeddings produced by the considered foundation models and minimizing the risk of losing information compared to other types of operation, such as sum or average.
Moreover, this operation allows to take advantage, in an equitable manner, of the variety of representations induced by the models pre-training on several large-scale datasets.

Following the MIL paradigm, we consider the aggregated embeddings $\{\hat{h}_{1}, ..., \hat{h}_{K}\}$ as a bag of instances, for which we seek an aggregation function to create the WSI-level embedding.
Due to the lack of a fine-grained labelization of the WSI (i.e., one label for the bag of instances), and the fact that all patches are not equally relevant for treatment response prediction, the aggregation function should be able to assign a highest importance to patches that most contribute to the prediction.
Inspired by the work of Ilse et al. \cite{Ilse2018AttentionbasedDM}, we define this aggregation function as a parametrized function, noted $g$:
\begin{equation}
\label{eq:attnpool1}
g(\hat{h}_{1}, ..., \hat{h}_{K}) = \sum_{i=1}^{K} \alpha_{i}\hat{h}_{i}
\end{equation}
with:
\begin{equation}
\label{eq:attnpool2}
\alpha_{i} = \frac{\exp \{ \mathbf{w}^{\top} \big{(} \tanh \big{(} \mathbf{V} \hat{h}_{i}^{\top}\big{)} \odot \mathrm{sigm} \big{(} \mathbf{U} \hat{h}_{i}^{\top} \big{)} \big{)} \} }{\sum_{i=1}^{K} \exp \{ \mathbf{w}^{\top} \big{(} \tanh \big{(} \mathbf{V} \hat{h}_{i}^{\top}\big{)} \odot \mathrm{sigm} \big{(} \mathbf{U} \hat{h}_{i}^{\top} \big{)} \big{)} \} }
\end{equation}
where $\mathbf{w}$, $\mathbf{U}$ and $\mathbf{V}$ are the learnable parameters.
This attention-based aggregation function is illustrated in Fig. \ref{fig:architecture} (see "WSI-level instances aggregation").

The WSI-level embedding is then forwarded to an MLP head to predict the treatment response.
It is worth mentioning that the WSI-level aggregation function $g$ and the MLP classifier, are the only trainable parts of our method.

\section{Experiments and Results}
\label{experiments}

\subsection{Data collection and processing}
\subsubsection{Dataset}
We conduct our experiments on a recently collected cohort of 152 patients from a French hospital (\textbf{the name of the hospital will be mentioned after the review process}), diagnosed with DLBCL between 2010 and 2023.
It includes 384 WSIs.
The collection has been reviewed by a group of senior and junior pathologists.
The DLBCL treatment response is evaluated by the Deauville score \cite{Meignan2009ReportOT}, which offers a finer scoring ranging from 1 to 5, based on the presence and the extent of the tumor.
For predicting the general outcome of the treatment, we group these scores into positive (1-3) and negative (4-5) responders.
We proceed to this grouping, as it is considered clinically relevant by experienced pathologists.
Among the 152 patients, 111 (73\%) are positive responders and 41 (27\%) are negative responders.
DLBCL affects multiple organs, and therefore, our dataset covers 38 different tissue types such as: lymph node, brain, mediastinum, soft tissue, bone, etc.

\subsubsection{Data processing}
For each WSI, tissue/background separation is done through Otsu thresholding segmentation on the HSV color space.
We extract 256$\times$256 non-overlapping RGB patches at $\times$20 magnification and exclude those with less than 50\% tissue coverage.

\subsection{Experimental settings}
\label{experiments:settings}
In our experiments we considered two foundation models and one standard ImageNet CNN as feature extractors.
The models corresponds to CONCH \cite{Lu2024AVF}, HIPT \cite{Chen2022ScalingVT} and ResNet-50 \cite{He2015DeepRL}.

CONCH \cite{Lu2024AVF} is a visual-language foundation model trained with SSL on 1.17 million patch–caption pairs, and we utilize its image encoder (without the language encoder).
Similarly, HIPT \cite{Chen2022ScalingVT} is also a ViT pre-trained with SSL on 104 million patches from 10K WSIs of The Cancer Genome Atlas (TCGA) database.
It is worth mentioning that the data used to pre-train these foundation models encompass multiple organs and cancer types.
Lastly, we consider ResNet-50 trained on ImageNet as the out-of-domain model.

At the patch-level aggregation function of our method, we compared our concatenation aggregation function to an attention-based one \cite{Ilse2018AttentionbasedDM}.

At the WSI-level aggregation function, we compared the proposed attention-based aggregation function (inspired by AB-MIL \cite{Ilse2018AttentionbasedDM}) to two other aggregation functions, namely TransMIL \cite{Shao2021TransMILTB} and DTFD-MIL \cite{Zhang2022DTFDMILDF}.
During the training stage of our method, we used binary cross-entropy as the loss function.

For the evaluation of all the variants of our method, we performed a $k$-fold cross-validation with $k=4$, and for each model, we report the mean and standard deviation of each metric across the four runs.

To overcome the class imbalance limitation of our dataset, we applied class weighting to the loss\footnote{calculated using the \textit{compute\_class\_weight} routine of the scikit-learn library} function, assigning a higher weight to the under-represented class.
\\
\\
\\

\subsection{Results}

\subsubsection{Contribution of foundation models over ImageNet pre-trained models}

\begin{figure}[t]
% 001 002 006 011
\centering
% trim={left bottom right top}
\includegraphics[trim={1cm 0cm 0cm 2cm},clip,width=\textwidth]{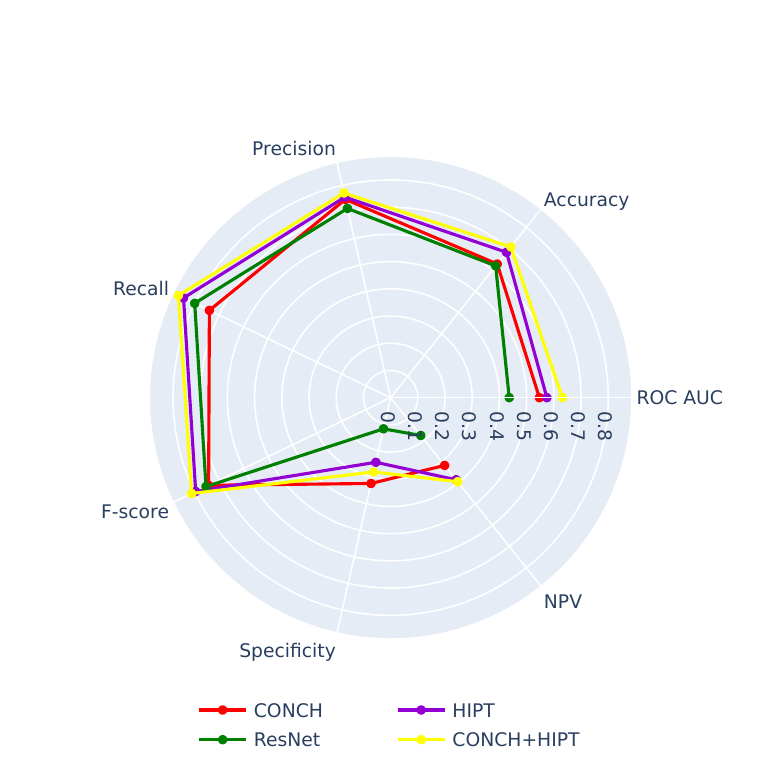}
\caption{Comparison of the performance of our treatment response prediction method by exploiting several foundation models and ResNet-50.}
\label{fig:numtiles}
\end{figure}

Fig. \ref{fig:numtiles} shows a radar chart summarizing the results obtained from our method by exploiting features extracted from several models, namely foundation models and ImageNet pre-trained ResNet-50.
The different metrics shown in the chart, indicate that when our method is trained on ResNet-based features, it obtains the lowest performance compared to foundation models-based features.
Moreover, even when the foundation models features are exploited individually (i.e., without patch-level features concatenation) we still observe better performance than ResNet (except for the recall, where we can observe  a slight improvement compared to CONCH).
One can also observe that aggregating two foundation models (CONCH+HIPT) provides overall better performance than using them individually.
This suggests that histopathology foundation models have stronger representation capabilities than ImageNet-based models.
However, low specificity results obtained across all feature extractors indicate that identifying negative responders is still challenging.
This low performance can be primarily attributed to the lack of representativeness of the negative class in our dataset.
Indeed, the application of class weighting in the loss function seems insufficient to overcome this limitation.
Nonetheless, when repeating the experiment without class weighting, we observed an overall diminished performance with an average decrease in specificity of 0.142.
Additionally, further experiments using techniques such as stochastic feature augmentation \cite{Li2021ASF} did not yield better results.

\subsubsection{Impact of patch-level aggregation function}

\begin{table}[t]
\caption{Comparison of aggregation functions performance. Highest value for each metric is highlighted in \textbf{bold}. We report each metric as \textit{mean(std)} over the $k$-fold cross-validation.}
\label{table:fusion}
\centering
\begin{tabular}{|l|l|l|l|}
\hline
\textit{Aggregation function}&  ROC AUC& F-score& Specificity \\
\hline
Concatenation& \textbf{0.632} (0.12)& \textbf{0.812} (0.053)& \textbf{0.28} (0.233) \\
Attention& 0.532 (0.052)& 0.621 (0.361)& 0.25 (0.413) \\
\hline
\end{tabular}
\end{table}

To study the impact of the proposed concatenation aggregation function, we compared our method with a variant in which we replaced the concatenation function by an attention-based aggregation function.
Comparison results are displayed in Table \ref{table:fusion}.
They clearly show the advantage of fully utilizing the image representations offered by foundation models compared to trying to learn an aggregation function for features selection.

\subsubsection{Impact of WSI-level aggregation function}

\begin{figure}[h]
% 003+012
\centering
% trim={left bottom right top}
\includegraphics[trim={1.5cm 1.5cm 0.5cm 2cm},clip,width=\textwidth]{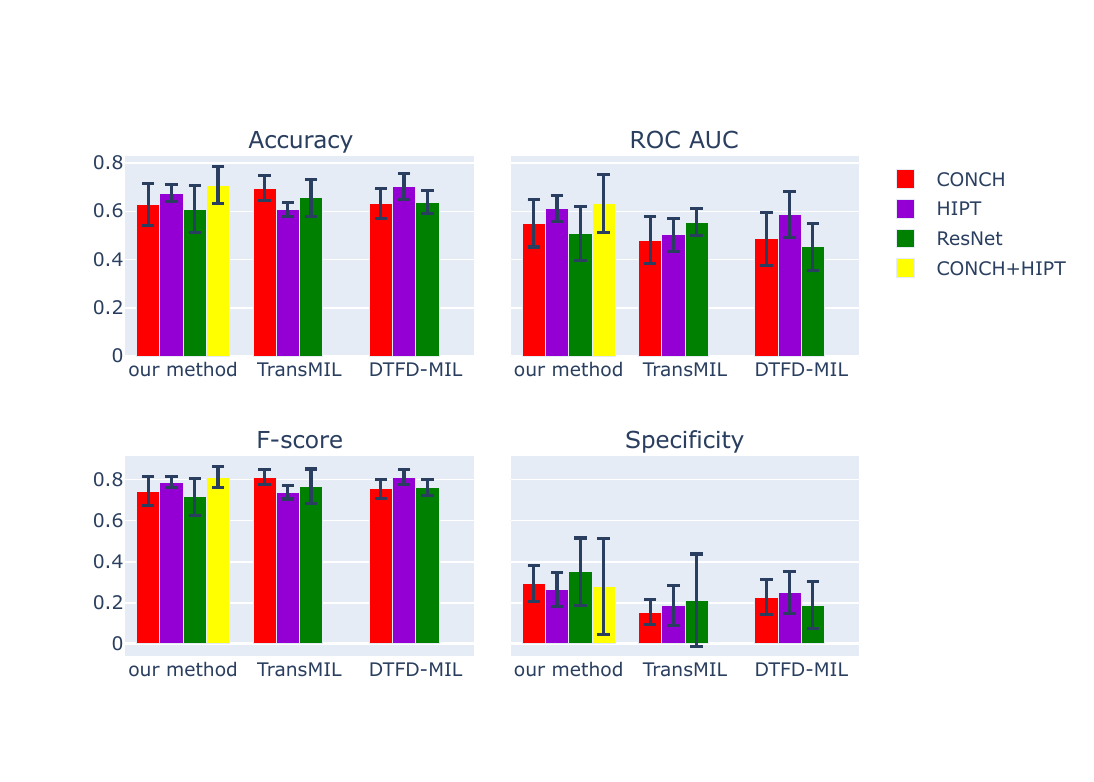}
\caption{Comparison of MIL methods with different feature extractors. Results are averaged from 4 runs and standard deviations are reported as error bars.}
\label{fig:001total}
\end{figure}

Fig. \ref{fig:001total} presents the comparative results between our method and two other MIL methods, namely TransMIL \cite{Shao2021TransMILTB} and DTFD-MIL \cite{Zhang2022DTFDMILDF}.
The two methods method have been trained by exploiting foundation models and ResNet-50 as feature extractors.
Owing to computational constraints for this experiment, all the methods were trained on 2,000 randomly sampled patches per WSI.
The bar plots show that our method achieved the highest performance overall metrics notably when exploiting the patch-level foundation models aggregation (see CONCH+HIPT in the figure).
Particularly, compared to the best results obtained by DTFD-MIL and TransMIL, our method provides a ROC AUC gain of 4.5\% and 7.6\% respectively and a specificity gain of 3.1\% and 6.7\% respectively.

\section{Conclusion}
Exploiting foundation models pre-trained on large-scale unlabeled datasets with self-supervised learning has been proven to be successful across multiple tasks in computational pathology \cite{Xu2024AWF,Lu2024AVF,Filiot2023ScalingSL,Chen2022ScalingVT}.
Complementary to these tasks, our proposed method together with the conducted experimental study on the new task of treatment response prediction demonstrated that these foundation models clearly have the potential to efficiently characterize histopathology images, and to produce more suited semantic representation than traditional ImageNet pre-trained models.
Moreover, the aggregation of these foundation models permits to enrich this semantic representation and improves the prediction capabilities.
In this sense, we encourage further investigations on aggregation methods and the exploitation of new foundation models proposed in the literature.

\begin{credits}
\subsubsection{\ackname}
This project has received funding from the \textit{French National Research Agency (ANR)} and the \textit{Protocole Région 2023}.
\subsubsection{\discintname}
The authors have no competing interests to declare that are relevant to the content of this article.
\end{credits}

\bibliographystyle{splncs04}
\bibliography{mybibliography}

\end{document}